# Human extraction and scene transition utilizing Mask R-CNN

Asati Minkesh
Master student in Electrical and Electronic Engineering
Tokai University
Kanagawa, Japan
minkeshasati@gmail.com

Kraisittipong Worranitta
Master student in Electrical and Electronic Engineering
Tokai University
Kanagawa, Japan
worranitta.mink@gmail.com

Miyachi Taizo
Professor in Electrical and Electronic Engineering/
ICT Center, Tokai University
Kanagawa, Japan
miyachi@keyaki.cc.u-tokai.ac.jp

*Abstract*—Object detection is a trendy branch of computer vision, especially on human recognition and pedestrian detection. Recognizing the complete body of a person has always been a difficult problem. Over the years, researchers proposed various methods, and recently, Mask R-CNN has made a breakthrough for instance segmentation. Based on Faster R-CNN, Mask R-CNN has been able to generate a segmentation mask for each instance. We propose an application to extracts multiple persons from images and videos for pleasant life scenes to grouping happy moments of people such as family or friends and a community for QOL (Quality Of Life). We likewise propose a methodology to put extracted images of persons into the new background. This enables a user to make a pleasant collection of happy facial expressions and actions of his/her family and friends in his/her life. Mask R-CNN detects all types of object masks from images. Then our algorithm considers only the target person and extracts a person only without obstacles, such as dogs in front of the person, and the user also can select multiple persons as their expectations. Our algorithm is effective for both an image and a video irrespective of the length of it. Our algorithm does not add any overhead to Mask R-CNN, running at 5 fps. We show examples of yoga-person in an image and a dancer in a dance-video frame. We hope our simple and effective approach would serve as a baseline for replacing the image background and help ease future research.

## I. INTRODUCTION

Human detection is a branch of computer vision and object recognition which appeared in various applications such as intelligent driving, and automated surveillance. The challenging part in object detection is to classify whatever in the picture and locate its location and pixels which we called it instance segmentation. There are many techniques in computer vision, the popular techniques are classification, semantic segmentation, object detection, and instance segmentation. Classification is to classify objects to a specific type of object (e.g., sky, balloon, car). Semantic segmentation is to understand the image in pixels and label with the class of its type of objects. Object detection is to indicate the spatial location of the object in the image and instance segmentation masks individual object with its location.

Over recent years, machine learning and deep learning have rapidly grown since Hinton's 2006 publication made a breakthrough in the artificial intelligence field and rebranded neural network to deep learning. In 2012, Alex Krizhevsky, Ilya Sutskever, and Geoff Hinton published ImageNet Classification with Deep Convolutional Neural Networks on the Large Scale Visual Recognition Challenge (LSVRC) contest which achieved a winning top-5 test error rate of 15.3%[1]. For object detection performance, combining region proposal with CNNs, Girshick introduced R-CNN, which is a scalable and straightforward object detection algorithm by a dominated object to its region proposal by selective search [2]. Considering the drawbacks of R-CNN, Girshick purposed another object detection algorithm named Fast R-CNN to improve the accuracy and speed of R-CNN and SPPnet [3]. Combining Fast-RCNN with Region Proposal Network (RPN), Faster R-CNN eliminates the selective search algorithm and lets the network learn the region proposals. Therefore, Faster R-CNN also capable of real-time object detection [4]. Extends Faster R-CNN, Mask R-CNN adding a branch for predicting an object mask parallel with the existing bounding-box recognition of Faster R-CNN for predicting a segmentation mask in a pixel-to-pixel manner [5].

The existing application for extract persons is Photoshop. Manually extract persons from an image is a hard job and even harder with video, we desire to create a simple but effective application with an automation system combining artificial intelligent technology. From now, extracting persons should not be manual work anymore.

We proposed an application to extracts multiple persons and put them into a new background image utilizing Mask R-CNN. We hope that our simple but effective approach could serve a baseline for replacing the image background and could be useful for future research.

## II. PROBLEM FOR EXTRACTION OF MULTIPLE PERSONS

Since the extracting person is a crucial task in image editing and Photoshop is the pioneer player in this field, that is why we identified the limitations of its latest version and tried to overcome those. Recently, Photoshop added a new feature called "select subject" which automatically select the subject or subjects in the image.

However, it still need to do some further tweaking separately. We would like to point out some of the limitations of Photoshop's automatic "select object" feature as below:

Problem1: Photoshop uses edge detection under the hood in its automatic object selection feature, so if there are multiple people or complex scenes, then it also considers the gap between them as objects and extracts it. So, we need to do further manual tweaking separately.

Problem2: Photoshop could not extract an individual human. If the image contains pet such as a dog or cat, it also includes them, but our application extract human only, and the user also can select multiple persons they would like to extract.

Problem3: Photoshop could not extract human from video; our algorithm can process video irrespective of the length of it. Also, extract those persons from a video and place them into the new background.

Our application does not have any of these limitations because we are using Mask R-CNN neural network trained on the COCO dataset for object detection. It can individually segment the objects and provide mask-information of 62 categories (e.g., horse, ball, human, chair) then our application select and extract only person objects from that image. Besides, we place extracted human-objects into the new background.

## III. RELATED WORK

### A. Existed work

Various current works concerned about scene understanding and object recognition, for example, in [7] and [8]. In [7], the authors have purposed a method to determine the basketball player who performs a given action among all the players in the game using Mask R-CNN for player detection and optical flow-based method for player measurement. According to [7], authors have to analyze the video frame by frame because instead of images, preparing the learning dataset of the video is more complex and each object have to be marked in all frame.

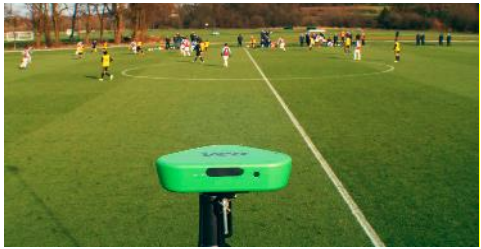

Fig. 1. Veo camera

Similar to [7], there is also research using Faster R-CNN inception model to capture the highlight of a football game. Preferentially with the exclusive option, there is also an artificially intelligent camera, for instance, Veo (See Fig.1), which can recognize the area of action in the football game, follow the ball position automatically and also capturing the highlight of the game. Veo camera features include soccer recording, coaching and also analysis. To make the soccer recording process uncomplicated and also be able to mark the highlight of the match.

Even now, artificial intelligence can understand the area of action in sports with and also capture the moment that human ability could not achieve.

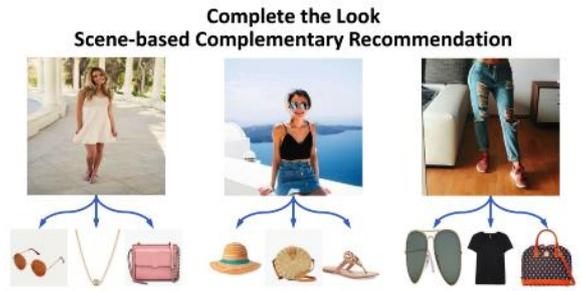

Fig. 2. Complete the look with scene understanding from [8]

On the other side, in W. Kang et al. showed the "Complete the look" (See Fig.2) application that uses scene understanding to recommend compatible products that have not been in the image for inspiration and shoppable product. It collected the context that could be useful for fashion such as outfit, indoor or outdoor, season, body type and the overall aesthetics of a room or places to efficiently predict products with visual search technology. In the advertising field, the application could be used for the clothes and accessories commercial in the future. As regards the research described above, our application for object detection and instance segmentation is also useful to pick target persons up for such recommendation.

### B. Object detection algorithm and dataset

#### 1) Mask R-CNN

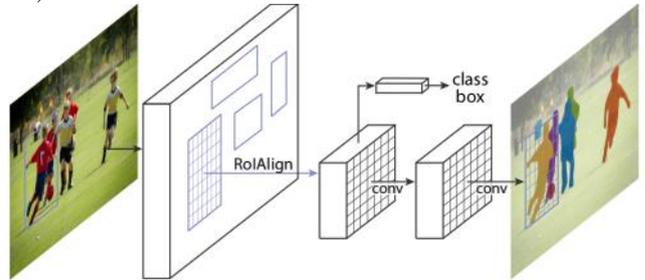

Fig. 3. Mask R-CNN framework for instance segmentation

The major goal of Mask R-CNN is to develop a framework for instance segmentation or image segmentation which is one of the computer vision tasks. Computer vision is seeking the techniques to help computers improve their understanding of digital images such as photos or videos.

Mask R-CNN is conceptually simple: Faster R-CNN has two outputs for each candidate object, a class label, and bounding box offset; to this Mask R-CNN add a third branch that outputs the object mask – which is a binary mask that indicates the pixels where the object is in the bounding box (See Fig.3). However, the additional mask output is distinct from the class and box outputs, requiring extraction of the much finer spatial layout of an object. Next, it includes a pixel-to-pixel alignment, which was the main missing piece of Faster R-CNN.

In general, Mask R-CNN is a spontaneous extension of Faster R-CNN, by adding a third branch of fully convolutional neural network for the object mask and instance segmentation and use RoI-Align instead of RoI-Pooling. Mask R-CNN algorithm is not only the state-of-the-art results but also achieved as a speedy system. Mask R-CNN also can easily be extended for human pose estimation by predicting K key points (e.g., left shoulder, right elbow) and by utilizing those key points, Mask R-CNN can be applied to detect instance-specific poses.

Therefore, Mask R-CNN can do works broadly and also can extend to more complex tasks. For the reader who is interested in Mask R-CNN, we recommended reading the reference [5].

*2) Comparison between Mask R-CNN and Faster R-CNN*

Semantic segmentation is a very challenging task especially extract persons from a video because it needs accuracy and a lot of datasets for training and also with a huge amount of datasets, it also required a lot of time consumed too. In this research, we create our own framework utilizing Mask R-CNN algorithm for object detection and then we extract those persons and replace them into a new background.

TABLE I. ROIALIGN VS. ROIPOOL FOR KEYPOINT DETECTION ON MINIVAL. THE BACKBONE IS RESNET-50-FPN FROM [5]

|  | AP | $AP_{50}$ | $AP_{75}$ | $AP^{bb}$ | $AP^{bb}_{50}$ | $AP^{bb}_{75}$ |
|---|---|---|---|---|---|---|
| *RoIPool* | 23.6 | 46.5 | 21.6 | 28.2 | 52.7 | 26.9 |
| *RoIAlign* | **30.9** | **51.8** | **32.1** | **34.0** | **55.3** | **36.4** |
|  | *+7.3* | *+5.3* | *+10.5* | *+5.8* | *+2.6* | *+9.5* |

Since Faster R-CNN and Mask R-CNN are the latest publication and quite popular among researcher on object recognition and instance segmentation field, we preferred Mask R-CNN rather than Faster R-CNN with these following reasons:

- Accuracy, Mask R-CNN replace RoI-Pooling that has been used in Faster R-CNN with RoI-Align which supports a very accurate instance segmentation and also pixel-to-pixel alignment.

- K keypoints detection, Mask R-CNN also can easily be extended for human pose estimation by predicting K keypoints (e.g., left shoulder, right elbow) and by utilizing those keypoints, Mask R-CNN can be applied to detect instance-specific poses.

Considering Table I, RoIAlign has shown magnificent improvement against RoIPool for keypoints detection which was very useful in person detection. That was the reason we rather used Mask R-CNN than Faster R-CNN.

As the development of learning datasets for both video and image analysis increases, scene understanding has a strong enthusiasm for that. Which it provides valuable information for various objectives such as security, marketing or for instance, customer behavior in the supermarket could be useful for the customer services and marketing department to optimizing the shopping performance and increasing the profit of the store. However, to achieve that kind of application, we need a significant and large number of learning datasets to create a model that generates an accurate result. The COCO dataset [6] has designed for object detection, and it has been frequently used in deep learning tasks because of its 80 object and 91 stuff categories, and the massive amount of labeled objects.

In this research, using Mask R-CNN that was trained on COCO dataset. We purposed a method to automatically extracts target person without other obstacles in both images or video and replace them to a new background.

*3) COCO dataset*

COCO dataset or Common Objects in Context is a dataset provided by Microsoft for object recognition of everyday scene images containing various object labeled using per-instance segmentations. (See Fig.4)

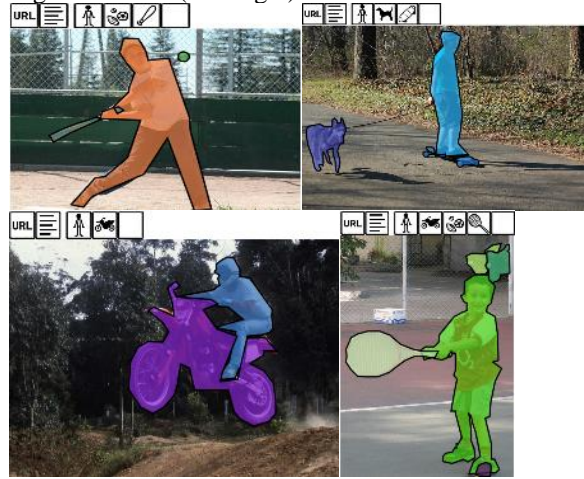

Fig. 4. COCO dataset that contains person example from [6]

The purpose of COCO dataset is for improving computer vision technology to understand the visual scene and recognizing the objects that appeared there, including the relationship of the object and semantic description of the scene. With 2.5 million labeled instances in 328k images, COCO dataset influenced enormous researcher works on instance spotting, category detection, and instance segmentation [6]. In COCO 2017 instance segmentation competition, the winning teams used Mark R-CNN framework with state of the art result.

Fig.4 shows the example of COCO dataset that contains persons, not only persons that can be detected but also the others object too. For instance, in the upper left image also contains baseball bat and baseball, the similarity with upper right side image that also contains dog and skateboard. We applied our application by separate only person object and replaced that person to a new background or scene.

IV. EXTRACTION METHOD OF MULTIPLE PERSONS

We proposed an application that detects and extracts only persons and place them with a new background. As an input, this application takes an image consists various objects and a background image, as an output, it gives a single image with an input background consisting of human objects extracted from an input image. It also works with video input, in that case, it will perform the same process frame by frame and after combining all frames it will give a video as an output. This application developed new useful functions for three problems in Section II.

Problem1: Our application uses Mask R-CNN for automatically extracts a bounding box that covers a target person and generates an ample contour of the person. Mask R-CNN uses Faster R-CNN architecture to detect the objects and FCN

(Fully Convolutional Network) to generate ample contour that covers the object, and then our algorithm extracts the target person object and place it into the new background. Problem2: Our application extracts a person only without obstacles, such as dogs in front of the person, and the user also can select multiple persons that users expect to extract.

Problem3: Our application is effective for both an image and a video irrespective of the length of it. Additionally, extract those persons from videos and place them with the new background.

Our application is capable of extracting single or multiple people from an image or video. Furthermore, users do not need to concern about the length of videos because if one video is shorter than others, then the length of the output video is equal to the most extended input video.

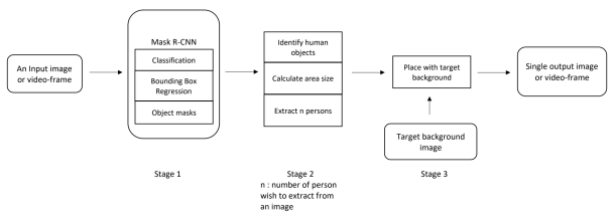

Fig. 5. Three stages of purposed method

Fig.5 shows an image of end to end application structure. This application takes two inputs and gives one output. The first input is an image or a video contains objects and scene, the second input is a target background image and the output is a single image or a video consisting of all extracted human objects from an input image placed into the target background.

This application has three stages. In the first stage, we do instance segmentation and generate the mask for each detected object for the input image using Mask R-CNN. In the second stage, we identify the human-object out of all types of objects and then select an expected number of human-object (based on their area size) which we need to extract from an input image. In the third stage, we extract the human-objects (selected in the second stage) and place them into the target background image.

We next describe how the three stages work in detail. Before feeding an input image into the first stage, we resized an input image and background image to the same size. In the beginning stage, we are using Mask R-CNN without any changes in their architecture for object detection (bounding box offset), object classification (object name) and mask generation which are pixels that belong to the object. In the next stage, we calculate the area of all human objects by using the bounding box offsets ($Y1, X1, Y2, X2$) with the following formula.

$$Area = (Y2 - Y1) * (X2 - X1)$$

$$[Rectangle\ Area\ =\ Height * Length] \qquad (1)$$

In order to be able to extract the main persons efficiently, we store the calculated area with respective person ids in descending order, then select the persons to extract according to number of person expected to extract from the input image (if we expect to extract n person from the input image then consider first n person ids). In the last stage, since an input image, generated mask, and background image are on the same size so we can compare pixel-by-pixel and replace the pixels of background image with human object pixels from an input image. For each selected human object in an input image, we replace pixel-value of the background image with the mask pixel-value where those pixels belong to human object.

## V. EXPERIMENTAL RESULT

### A. Hardware environment

In this project, we used an i5 processor with 6-cores, 32-GB RAM, 3-TB ROM, and a graphics card with 12-GB memory. We were able to process around 3-frames per second when 1-image were feeding into the network in parallel, and 5-frames per second when 2-images were feeding into the net in parallel.

### B. Implementation

In the first stage of our implementation, we used the open-source implementation of Mask R-CNN using Python, Keras, and TensorFlow [15]. It is based on Feature Pyramid Network (FPN) and Resnet101 backboned. Most of this implementation follows Mask R-CNN, but there are some cases where they deviated in favor of code simplicity and generalization. There are mainly 3 cases that differences. First, they resized all images to the same size to support training multiple images per batch. Second, they ignored bounding boxes that come with the dataset and generate them on the fly to support training on multiple datasets because some dataset provides bounding boxes and some provide masks only. This also made it easy to apply image augmentation that would be harder to apply to the bounding boxes, such as image rotation. Third, they used a lower learning rate instead of using 0.02 that was used in the original paper. Because they found it to be too high, that often causes the weights to explode, especially when using a small batch size.

The second stage and third stage are implemented using Python and its libraries for extracting and placing the extracted objects into a new (target) background. We structured the code in a way to be able to extract single or multiple people out of all detected human objects from the image.

In the case of video input, the video will be divided into frames and then application iterate all 3 stages on each frame. After all iterations, all frames will be combined in a single video with the same frame rate as input video and this video will be our output.

### C. Results

This application principle is to create a framework that extracts and replace person to the new background for a furthermore application. For instance, video editing, a scene change in advertising or even security camera to observe the behavior of the suspicious person in a chaotic environment and also to see the dance move clearly and the exercise such as yoga or aerobic move more evident than before with the original background.

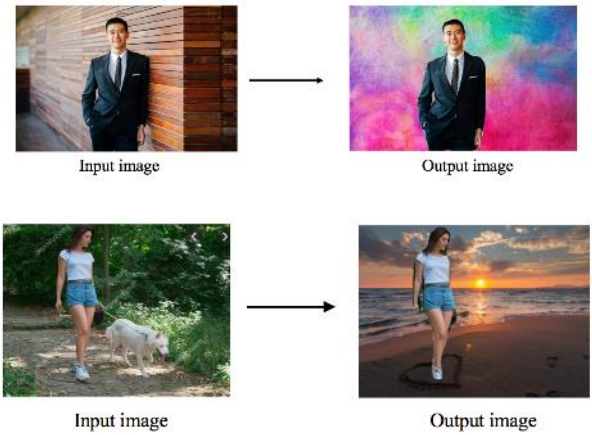

Fig. 6. Example of the result in a single person

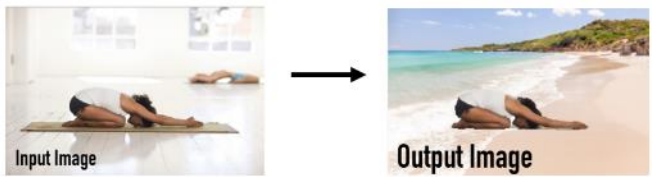

Fig. 7. Example of the result with yoga pose

We have shown four types of examples. Firstly, Fig.6 and Fig.7 show an example of extracting a single person. In the image above we placed that person to a new fantasy background which might be useful for modifying the images with a chaotic background to a more delightful background. Similarly, with the image below, we extracted the woman without any obstacle such as a dog in the image.

Further, the yoga pose of the woman in a studio in Fig.7, we placed that person to the beach background. A user can easily make a poster of "Beach Yoga" and it also could be applied to the cleaner background to observe the pose which could apply for extracting the pose of the others exercise too.

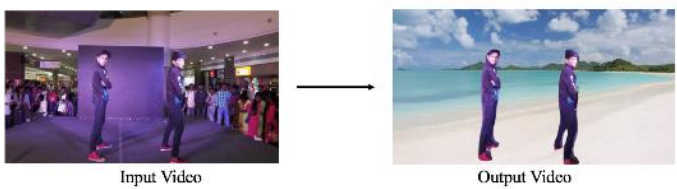

Fig. 8. Example of the result in multiple persons

Secondly, as shown in Fig.8, our application also capable of extracting multiple people simultaneously at the same time, which extracts the persons from a crowned stage to a beach scene. In this example is also show that even the scene is crowded and they are a lot of human object in the image but our application calculated only target humans according to the significant area of them.

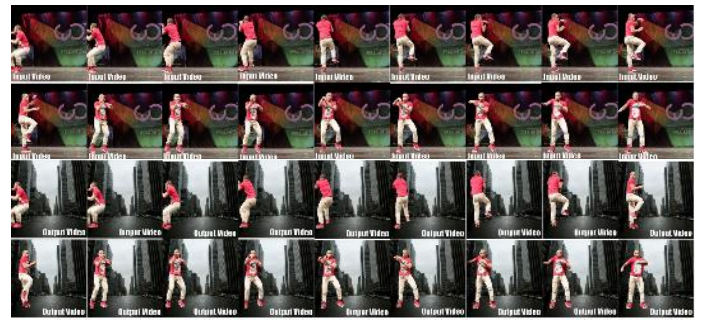

Fig. 9. Example of the result with the dance move

Thirdly, an example of the possible application shows in Fig.9 which replaced the dancing person to a new scene which easier to identify the movement of that person smoothly. In Fig.9 also shows the example of extracting a person from video frames to a new scene frame by frame.

Lastly, in Fig.10 shows the example of a happy family with movie scenes such as The Sound of Music. We extract the family from the original photo and place them to a new pleasant background and also famous places.

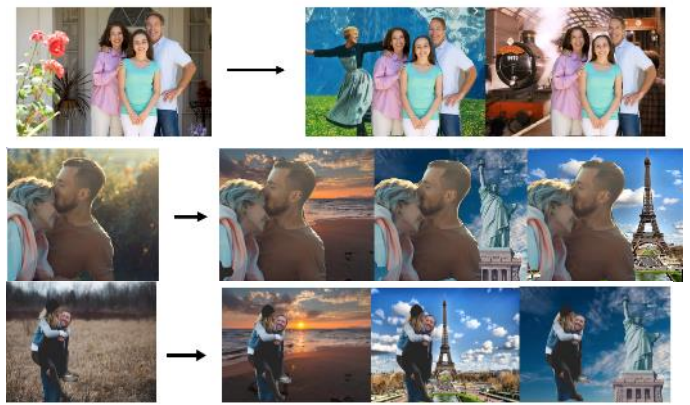

Fig. 10. Example of the result with movie scenes and famous places

Moreover, this result examples could apply with tours and travel agencies advertisement campaigns. In the tourism business, marketing has to work on creative and interesting campaigns to attracts more customers. Artificial intelligence could be powerful solutions for digital advertising especially on content creation such as Fig.10 has shown.

The demonstration of customers in a beautiful scene or places could bring imagination into vision. With this application, a travel agency could deliver unique and pleasant experiences and the customer can try on any places they interested in.

*D. Discussion*

We believe that our application can significantly benefit for further purpose. We could extract only a human object or a human with another object such as animals. We have an idea of applying the Mask R-CNN algorithm with a warning system. For instance, we detect the human objects from the beach scene and analyzing the information in real-time for detecting the persons who might get drowned and calling for help.

The accuracy of human detection is still an ongoing project. Even though Mask R-CNN has a great result on the accuracy and the adequate framework for instance-level tasks, we believed that artificial intelligence researcher and also Mask R-CNN developer themselves would continue the effort for improving object detection and instance segmentation. Correspondingly, our future work shall have to consider those improvements conform to our application for more precise detection and contribute to the valuable idea of a new application.

## VI. CONCLUSION

In this paper, we purposed human extraction and scene transition using Mask R-CNN. The main objective of this research is to build a new application of extracting persons from an image or video and place them into a new background using an artificial intelligence approach. Although It can process 5 frames per second but wound not be sufficient for real-time application. Because our application is dependent on Mask R-CNN for object detection that is why we could not improve the accuracy in human detection. To improve the accuracy we need to design a new network and train it on a large dataset of human images then we would be able to make a system for real-time application. Apart from that, one possible future work is to modify the application to support multiple input images or videos in which users also could select the number of persons they would like to extract from each input image and merge those persons to the new background.

ACKNOWLEDGMENT

We want to thank JICA (Japan International Cooperation Agency) for the support and also Matterport for open-sourcing the implementation of Mask R-CNN based on TensorFlow.